\documentclass[letterpaper, 10 pt, conference]{ieeeconf}

\IEEEoverridecommandlockouts
\usepackage{graphics}    
\usepackage{epsfig}      
\usepackage{times}
\usepackage{amsmath}
\usepackage{amssymb}

\usepackage{cite}

\makeatletter
\let\NAT@parse\undefined
\makeatother

\usepackage{xcolor}
\usepackage{booktabs}
\usepackage{tabularx}
\usepackage{ulem} 
\usepackage{algorithm}
\usepackage{algpseudocode}
\usepackage{graphicx}
\usepackage[font=small,labelfont=bf]{caption}
\usepackage[inkscapelatex=false]{svg}
\usepackage[export]{adjustbox}
\usepackage{siunitx}
\let\labelindent\relax \usepackage{enumitem}
\usepackage{enumitem}
\usepackage{cuted}
\usepackage{capt-of}   
\usepackage{hyperref}

\definecolor{darkyellow}{rgb}{0.8, 0.8, 0}
\newcommand{\methodname}{CoDex}

\title{\LARGE \bf
CoDex: Learning Compositional Dexterous Functional Manipulation without Demonstrations
}

\author{
  Bowen Jiang, William Painter Reger, Roberto Martín-Martín
  \thanks{All authors are with Robot Interactive Intelligence Lab (RobIn), University of Texas at Austin}
  \thanks{This work was supported in part by a Meta research gift. Any opinions, findings, and conclusions expressed in this material are those of the authors and do not necessarily reflect the views of the sponsors.}
}

\begin{document}
\maketitle


\thispagestyle{empty}
\pagestyle{empty}


\begin{abstract}

    In this work, we study \textit{Compositional Dexterous Functional Object Manipulation} (CD-FOM): tasks such as aiming and actuating a spray bottle on a plant or a glue gun on wood, which require both actuating an object's internal mechanism and controlling its pose to apply the object's function to the environment.
    These tasks pose significant challenges for robots due to the demanding integration of semantic understanding ---of the object's function, actuation mode, and application area--- with intricate physical dexterity ---to manage grasp stability, movement trajectory, and actuation. 
    We introduce \methodname{}, a zero-demonstration framework that autonomously discovers CD-FOM manipulation strategies. 
    \methodname{} uses vision–language models (VLMs) to infer semantic constraints from the task and scene. 
    These constraints guide analytic constrained optimization to generate a short list of functional grasp candidates that can be efficiently refined with reinforcement learning to generate full grasp–move–actuate policies transferrable from simulation to the real world.
    We evaluate \methodname{} on a 7-DoF robot arm with a 16-DoF multi-fingered hand across six CD-FOM tasks involving previously unseen objects with internal mechanisms (spray bottles, hot glue guns, air dusters, flashlights, pepper grinders) and their application to unseen target objects, showcasing its ability to autonomously discover and execute complex, physically viable dexterous behaviors without human demonstrations. 
    More information at \url{https://robin-lab.cs.utexas.edu/CoDex/}.
\end{abstract}

\section{Introduction}
\label{s:intro}

Imagine a robot tasked with spraying a plant: it must grasp the bottle stably, aim it toward the leaves, and squeeze the trigger to release the spray, all in a coordinated sequence. 
This type of task is a form of Functional Object Manipulation (FOM)~\cite{paulius2016functional,srinivasan2024dexmots,huang2025fungrasp,aburub2025eigengrasp,agarwal2023dexfuncgrasp}, where the robot actuates an object's internal degrees of freedom (e.g., trigger, button, lever) while controlling its external DoF (the object pose) to apply the function to a target region~\cite{Li2022DexReview, An2025DexSurveyIL}. 
We refer to these problems as \textbf{Compositional Dexterous Functional Object Manipulation (CD-FOM)}. 
Because they require coordinating internal actuation with precise object motion, CD-FOM tasks remain a major challenge in robotics.


CD-FOM tasks require bridging semantic understanding and physical dexterity. 
The robot must interpret the task context—what the object is for, where it should be actuated, and where its effect should be applied—while executing precise physical interactions such as stable functional grasps, coordinated arm–hand motion, and controlled force application. 
This calls for guiding physical skill learning with semantic reasoning.

\begin{figure}
    \centering
    \includegraphics[width=\linewidth]{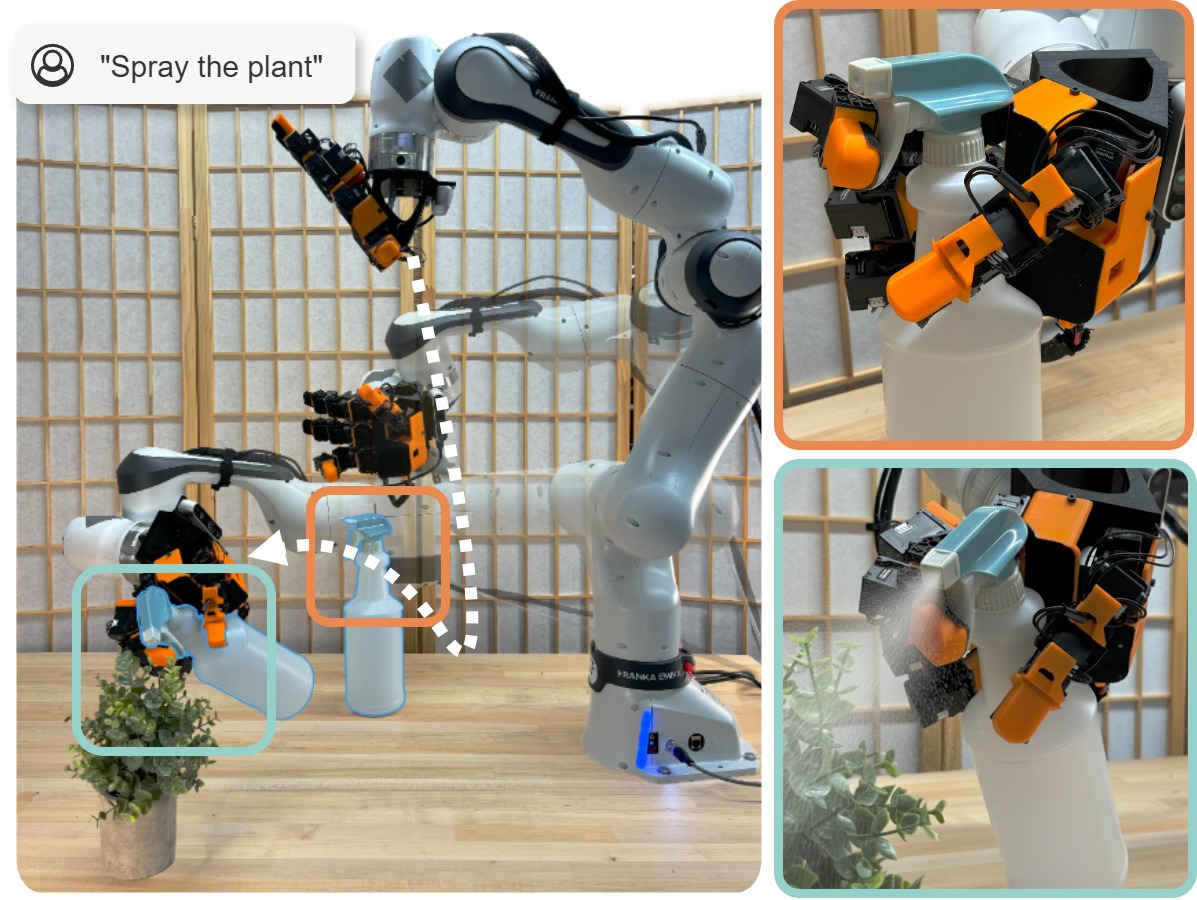}
    \caption{\textbf{Robot executing a Compositional Dexterous Functional Object Manipulation (spraying water on a plant) using \methodname{}}. The robot autonomously performs a task-aware functional grasp, repositions the spray bottle, and actuates the trigger to spray mist, all without any human demonstration. \methodname{} guides these tasks by bridging high-level semantic understanding through a VLM with low-level physical dexterity via constrained optimization and reinforcement learning.}
    \label{fig:teaser}
    \vspace{-1.51em}
\end{figure}

General object manipulation methods fall short for CD-FOM.
Learning from Demonstration methods acquire dexterity from expert teachers, while semantic understanding is implicitly encoded in their behavior~\cite{zhao2023aloha,mandlekar2022matters,iyer2024open,wang2024dexcap}.
However, learning the correlation between semantics and dexterity from demonstrations is difficult because it requires large amounts of data collected through teleoperation of complex multi-fingered hands to actuate objects with internal mechanisms~\cite{handa2020dexpilot,wang2024dexcap,iyer2024open,Levine2016}.
Recent imitation-from-human-videos methods remove the need for labor-intensive teleoperation, but instead require overcoming human–robot morphological differences with limited object-specific strategies~\cite{bahl2023affordances,An2025DexSurveyIL,bahety2024screwmimic,bahety2025safemimic}.
Alternatively, optimization-based approaches such as reinforcement learning~\cite{sutton1998reinforcement,kaelbling1996reinforcement} and analytical grasp synthesis~\cite{shimoga1996robot,miller2000graspit,berenson2008grasp,morrison2018closing,li2023frogger} achieve the physical dexterity required for CD-FOM without demonstrations. 
However, their lack of semantic understanding requires external object-specific guidance in the form of reward design and optimization objectives, limiting their applicability and autonomy~\cite{Charlesworth2021,agarwal2023dexfuncgrasp,srinivasan2024dexmots,SpringGrasp2024,zhang2025dextog}.
General semantic understanding can instead be obtained from large-scale pre-trained models such as Vision-Language Models (VLMs)~\cite{Ma2024VLASurvey,DexVLA2024,RoboMamba2024,nasiriany2024pivot,huang2024rekep}. 
However, early integrations of VLMs into robotic systems~\cite{huang2024rekep,nasiriany2024pivot} revealed limitations in geometric and embodied understanding, restricting their guidance to coarse, abstract levels that are insufficient for the intricate, coordinated hand–arm motions required in CD-FOM~\cite{Ma2024VLASurvey,DexVLA2024,RoboMamba2024}.

In this work we introduce \textbf{\methodname{}}, a framework that bridges semantic understanding and physical dexterity for CD-FOM through the use of \textit{semantic constraints}. 
We define semantic constraints as a set of geometric and spatial conditions derived from an object's function and the overall task goal. \methodname{} integrates a VLM into an iterative refinement procedure to achieve zero-demonstration semantic understanding, interpreting the task and generating two types of semantic constraints: local (e.g., where to press a trigger and in which direction) and global (e.g., where to aim a nozzle).
These constraints guide a two-stage learning pipeline. 
First, analytic constrained optimization generates functionally valid grasp candidates. 
Second, reinforcement learning refines these candidates into a full grasp–move–actuate policy.

We demonstrate the capabilities of \methodname{} to operate six previously unseen objects for different CD-FOM tasks, controlling a 7-DoF robot arm equipped with a 16-DoF multi-fingered hand and achieving 73\% average combined success rate. Our experiments validate that the VLM-generated semantic constraints are crucial to this performance. A human participant study demonstrates that our method for determining global constraints produces significantly more appropriate poses than prior VLM-based approaches. Furthermore, we show that our final policy learning stage is critical for achieving physical dexterity, improving functional success by over 40\% when compared to using analytical grasps combined with the same VLM-generated constraints.

\section{Related Work}
\label{sec:related_work}

\methodname{} bridges semantic understanding and physical dexterity for compositional dexterous functional object manipulation (CD-FOM). We position our work relative to three key research areas.

\begin{figure*}[t]
    \centering
    \includegraphics[width=\textwidth]{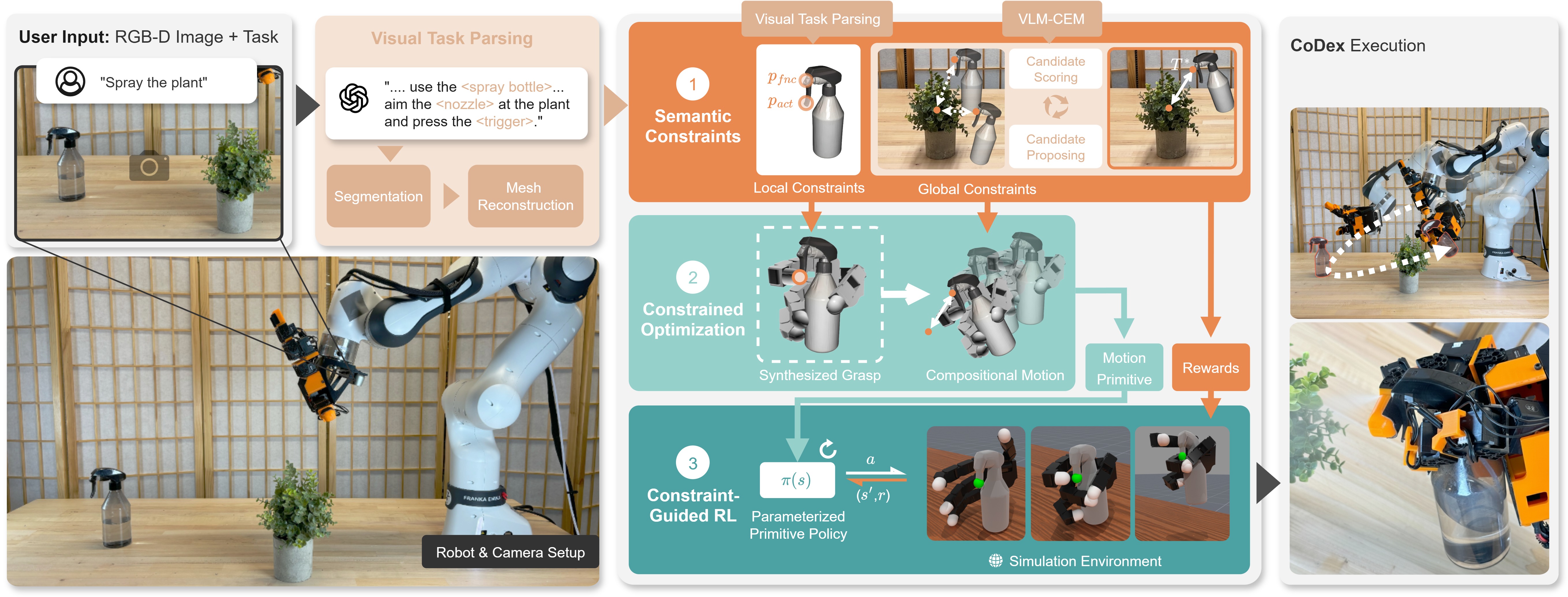}
     \caption{Overview of the \methodname{} pipeline. \methodname{} bridges high-level VLM understanding and low-level dexterity by translating abstract VLM outputs into concrete semantic constraints that guide a two-stage policy learning process.
    (1) \textbf{VLM-Generated Semantic Constraints.} First, a VLM interprets the user's input to generate local constraints (key interaction points like the actuation point and function point) and a global constraint (the final object pose).
    (2) \textbf{Constrained Optimization.} In the first learning phase, these constraints are enforced through analytic constrained optimization to synthesize a diverse set of motion trajectories, each of which includes a task-aware functional grasp.
    (3) \textbf{RL Policy Training.} In the second phase, these motion trajectories initialize a constraint-guided RL process, which uses the same semantic constraints as a reward function and learns the complete grasp-move-actuate policy.}
    \label{fig:method_overview}
\end{figure*}

\textbf{Semantic Understanding via Vision-Language Models.} 
Vision–Language Models (VLMs) provide strong zero-shot semantic understanding for robotic tasks, allowing robots to interpret goals from language and visual context~\cite{Ma2024VLASurvey, huang2024rekep, nasiriany2024pivot, DexVLA2024, RoboMamba2024}.
However, VLM outputs are typically abstract and lack the geometric precision required for dexterous manipulation~\cite{Ma2024VLASurvey, DexVLA2024, RoboMamba2024}.
Recent VLM-based manipulation systems~\cite{DexVLA2024, RoboMamba2024, RoboDexVLM2025} show promise but either require extensive training data or remain limited to coarse manipulations. ReKep~\cite{huang2024rekep} uses VLMs to generate keypoint constraints for manipulation, while PIVOT~\cite{nasiriany2024pivot} employs iterative visual prompting to refine robot actions.
\methodname{} leverages VLMs to generate concrete semantic constraints, both local (actuation and function points) and global (target poses), that directly inform constrained optimization and policy learning, effectively translating abstract VLM knowledge into CD-FOM policies without requiring task-specific training data.

\textbf{Functional Object Grasping and Physical Dexterity.}
In task-oriented grasping, the aim is to select a grasp that not only stabilizes the object but also facilitates the intended function~\cite{agarwal2023dexfuncgrasp, aburub2025eigengrasp, zhang2025dextog, li2025sayfuncgrasp, brahmbhatt2019contactgrasp, sundermeyer2021contactgraspnet, SpringGrasp2024}. 
Several methods focus on optimizing contacts to achieve stable grasps, often predicting force-closure metrics like the Ferrari-Canny~\cite{Ferrari1992, brahmbhatt2019contactgrasp, sundermeyer2021contactgraspnet, SpringGrasp2024}, but cannot be applied to grasps that enable the actuation of objects' internal degrees of freedom. 
Recent analytical strategies~\cite{aburub2025eigengrasp,li2025sayfuncgrasp,zhang2025dextog} consider internal degrees of freedom but they do not compose them with post-grasping trajectories to actuate the object at the right location to achieve a task.
All these methods aim to provide a grasp synthesis solution that generates a successful grasp from images to be executed by a predefined controller, which can lead to failures. 
Recently, some methods have integrated a simulator with a model of the specific object into the loop for online improvement of grasping strategies using reinforcement learning or exploiting the simulator's differentiability for optimization~\cite{wang2020manipulation, Weng2023, srinivasan2024dexmots, SpringGrasp2024, Sundaralingam2024DiffRolling}.
While demonstrating better performance, this strategy requires manual annotation of the rewards and has yet to be extended to complex objects with internal degrees of freedom (DoF) and post-grasping motion.
Moreover, all previous methods improve grasp stability and/or functionality, but treat grasping as an isolated problem, missing the opportunity to reason about it in conjunction with subsequent motion to enhance dynamics and stability during actuation.


\textbf{Compositional Dexterous Functional Object Manipulation.} Often, successful tool use demands composing the control of both in-hand adjustments and whole-arm extrinsic motions~\cite{An2025DexSurveyIL, Charlesworth2021, Li2022DexReview, Zhou2024BiDexHD}, but most existing works on dexterous manipulation focus on one or the other. 
For instance, significant research addresses in-hand manipulation, focusing on fine finger coordination for tasks like object reorientation~\cite{agarwal2023dexfuncgrasp, Sundaralingam2024DiffRolling, Charlesworth2021, Levine2016}, or rotating caps~\cite{srinivasan2024dexmots, VisuomotorDiffusion2025, KineSoft2025}, 
without consideration for full arm motion to control the object's overall trajectory for a task. 
Other works tackle extrinsic manipulation, where a grasped tool interacts with the environment, such as hammering or shoveling~\cite{agarwal2023dexfuncgrasp,lin2024hato,fang2020learning} on objects without internal degrees of freedom. 
Closer to ours, \cite{agarwal2023dexfuncgrasp} provides a composed solution for arm and multi-fingered hand motion, but that method requires human demonstrations and focuses on optimizing grasp stability to resist forces from subsequent arm motion, failing to actuate the object's internal degrees of freedom.
In contrast, \methodname{} holistically addresses the entire grasp-move-and-actuate problem, generating a composed solution that actuates both the object's internal and external degrees of freedom.

\section{\methodname{}: Compositional Dexterous
Functional Object Manipulation}
\label{sec:method}

    

\begin{figure}[t!]
\centering
\includegraphics[width=\linewidth]{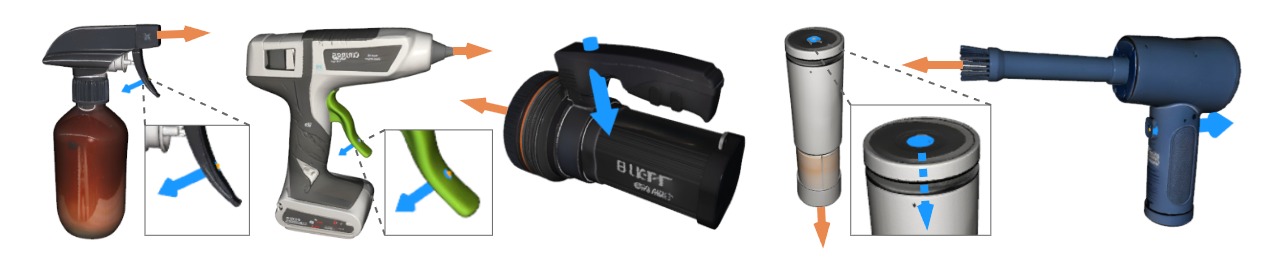}
\caption{Reconstructed objects with their VLM-identified \textbf{local semantic constraints}. The generation process combines semantic and visual information from VLMs (see Sec.~\ref{sec:vlm_guidance}) to infer the
actuation point, $p_{\text{act}}$, (blue arrow start) and function point, $p_{\text{fnc}}$, (orange arrow start). The actuation direction $d_{\text{act}}$ and function direction $d_{\text{fnc}}$ are parallel to the surface normal at the actuation and function points, pointing into the object (dotted line indicates \textit{inside the object}). These semantic constraints guide the grasp and motion optimization and training processes.}
\label{fig:vlm_actuation}
\vspace{-1em}
\end{figure}
 
\methodname{} bridges semantic understanding and physical dexterity by translating VLM reasoning into explicit geometric task constraints that can be enforced by optimization and reinforcement learning.
As illustrated in Fig.~\ref{fig:method_overview}, given a language task description $\mathcal{L}$ and an RGB-D scene observation $\mathit{I}$, our pipeline sequentially executes two stages: VLM-Generated Semantic Constraints and Constraint-Guided Policy Training. Constraint-guided Policy Training can be further divided into two sub-stages: Constrained Optimization and Constraint-Guided RL.  Below, we detail each stage.

\subsection{VLM-Generated Semantic Constraints}
\label{sec:vlm_guidance}

Given the input pair $(\mathcal{L}, I)$, this stage uses VLMs to generate semantic task constraints that guide policy learning. These constraints include: 1) \textbf{local constraints} (the actuation point $p_{\text{act}}$ and function point $p_{\text{fnc}}$ on the object, and the target point $p_{\text{tgt}}$ on the environment) and 2) \textbf{global constraints} (the functional object's target 6D pose $T^*$).

The process starts with \textit{Visual Task Parsing}. The functional object is identified from $(\mathcal{L}, I)$ using open-vocabulary segmentation (LangSAM). The segmented functional object's 3D mesh $\mathcal{M}$ is then constructed using a shape reconstruction and completion method (\mbox{Tripo~\cite{tochilkin2024triposr}}). The reconstructed meshes are scaled to match the heights measured in the scene Point Clouds un-projected from the RGB-D images. 

\subsubsection{Local Semantic Constraints}

Next, to derive the local semantic constraints, \methodname{} queries a VLM to generate text descriptions of key task-relevant interaction points (e.g., ``trigger'', ``nozzle'') that are then linked to 2D image pixels using MoLMo~\cite{deitke2024molmo} and unprojected into 3D by aligning the image and the reconstructed mesh using an image-based tracker (FoundationPose~\cite{wen2024foundationpose}) and matching 2D to 3D mesh points. This generates the \textbf{Actuation Point} $p_{\text{act}}$ and \textbf{Function Point} $p_{\text{fnc}}$. \methodname{} assumes that the actuation direction, $d_{\text{act}}$, is parallel to the surface normal at $p_{\text{act}}$, pointing into the object. Fig.~\ref{fig:vlm_actuation} depicts example results of this process.

\subsubsection{Global Semantic Constraints}


Finally, \methodname{} derives \textbf{global semantic constraints} (the goal pose $T^*$) through \textit{VLM-Guided Cross-Entropy Method} (VLM-CEM), an algorithm inspired by~\cite{nasiriany2024pivot}. VLM-CEM leverages the VLM’s reasoning to drive an iterative pose search: at each round, the VLM is prompted with the history $\mathcal H$ of previously scored candidate poses and proposes $K$ new candidate poses expected to score higher. We render these candidates (see Fig.~\ref{fig:vlm_cem_more_vis_app}), score them with the VLM, append them to $\mathcal H$, keep only an elite subset, and repeat for $N$ rounds. As detailed in Algorithm~\ref{alg:vlm-cem}, this iterative proposal-and-evaluation loop allows the VLM to perform an implicit optimization, converging on a pose that is both functionally correct and physically grounded.

\begin{algorithm}[]
\caption{VLM-CEM}
\label{alg:vlm-cem}
\begin{algorithmic}[1]
\Require Function Point $p_{\mathrm{fnc}}$, Target Location $p_{\mathrm{tgt}}$, iterations $N{=}6$, candidates $K{=}10$
\Ensure goal pose $T^*$
\State $T_0 \gets \textsc{AnchorInit}(p_{\mathrm{fnc}},\, p_{\mathrm{tgt}})$
\State $s_0 \gets \textsc{VLMScore}(\textsc{Render}(T_0))$
\State $\mathcal H \gets \{(T_0, s_0)\}$
\Statex
\For{$i = 1$ \textbf{to} $N$}
  \State $\mathcal P \gets \textsc{VLMPropose}(\mathcal H,\, K)$
  \For{\textbf{each } $T \in \mathcal P$}
    \State $s \gets \textsc{VLMScore}(\textsc{Render}(T))$
    \State $\mathcal H \gets \mathcal H \cup \{(T, s)\}$
  \EndFor
  \State $\mathcal H \gets \textsc{TopR}(\mathcal H,\, \rho)$ \Comment keep top-$\rho$ elites (optional)
\EndFor
\State \Return $T^* \gets \arg\max_{(T,s)\in \mathcal H}\, s$
\end{algorithmic}
\end{algorithm}


\begin{figure}[t!]
    \centering
    \includegraphics[width=\linewidth]{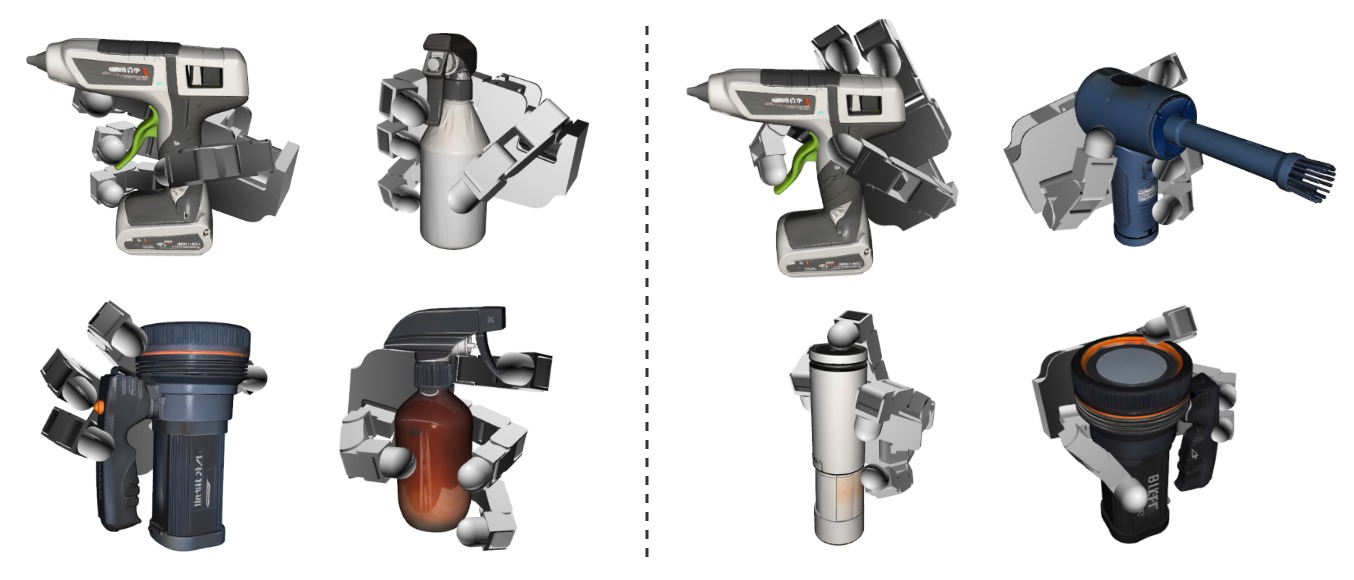} 
    \caption{Human-like (\textit{left}) and robot-specific (\textit{right}) examples of initial functional grasp candidates. Our analytic constrained optimization synthesizes functionally valid human-like and robot-specific grasps allowing \methodname{}  to exploit the hand's full morphology instead of restricting it to the human grasps that can be obtained with imitation learning.}
    \label{fig:grasp_examples} 
    \vspace{-1em}
\end{figure}

\begin{figure}[t]
\centering
\includegraphics[width=\linewidth]{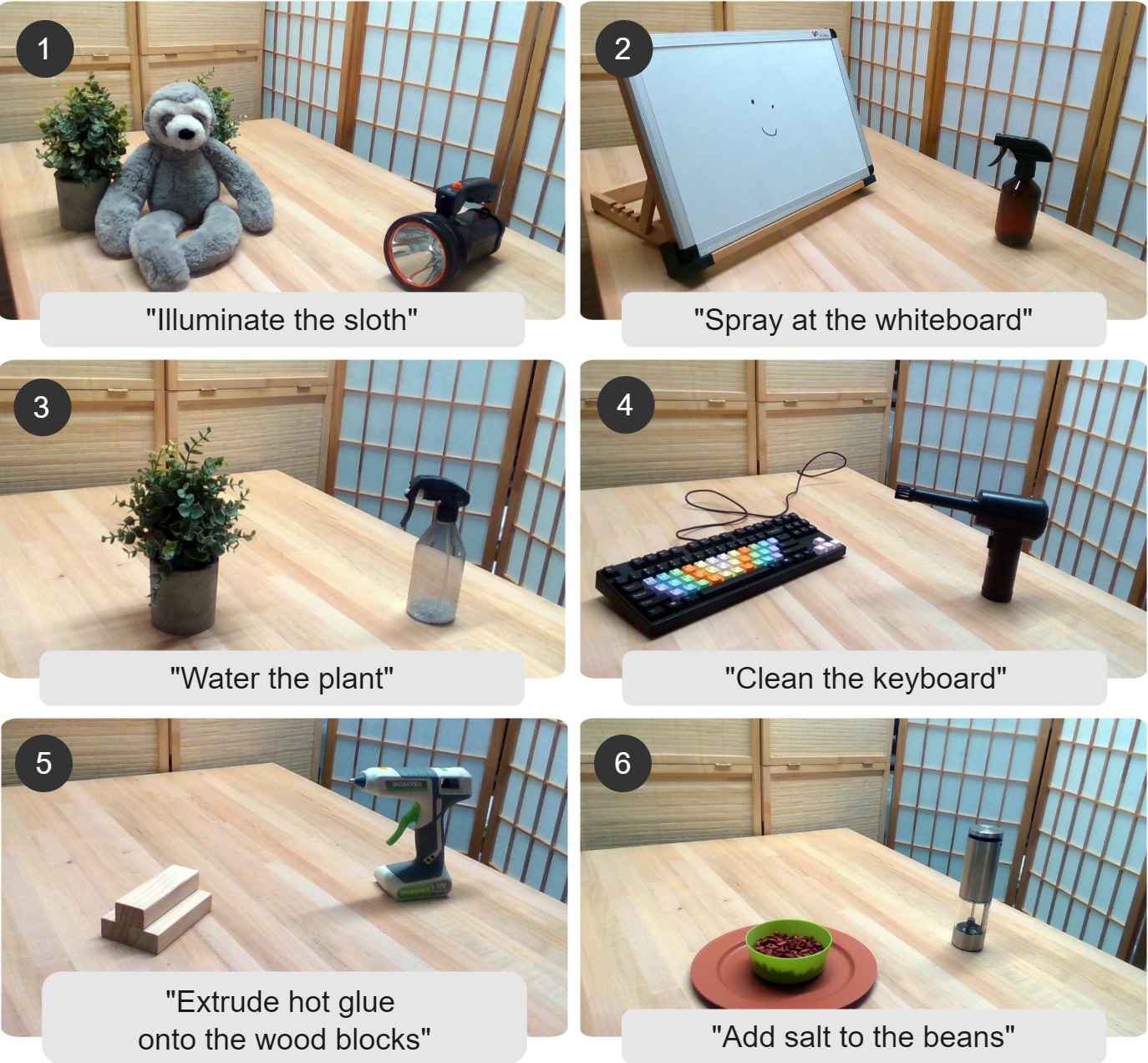}
\caption{Six functional object manipulation tasks in our experiments. They require combining local manipulation of functional objects with internal DoF (flashlight, board spray, water spray, air blower, hot glue gun, and salt grinder) with their global motion in the scene.}
\label{fig:tasks_objects}
\vspace{-1em}
\end{figure}

\subsection{Constraint-Guided Policy Training}
\label{sec:grasp_generation}


Using the semantic constraints generated in the previous stage, this stage learns a policy $\pi$ for the complete grasp-move-actuate sequence. Rather than learning dexterous manipulation from scratch, \methodname{} first uses analytic constrained optimization to synthesize a diverse set of functionally valid grasp candidates. 
These candidates provide structured initialization for reinforcement learning, dramatically reducing the exploration required to learn the full grasp–move–actuate behavior.

\subsubsection{Analytic Constrained Optimization}


This phase translates the VLM-generated local semantic constraints into concrete mathematical objectives for grasp synthesis. We first sample initial hand configurations $q_0$ from $\mathcal{Q}$---the valid joint space---using inverse kinematics, biasing a finger to be near the actuation point $p_{\text{act}}$. These samples are then refined via constrained optimization. The local constraints ($p_{\text{act}}, d_{\text{act}}$) directly inform the functional terms in Eq.~\ref{eq:grasp_opt_appendix_full}: the \textit{Actuation Pt Proximity} term ensures a fingertip is within a distance $\delta_{\text{dist}}$ of $p_{\text{act}}$, while the \textit{Actuation Alignment} term ensures the fingerpad normal aligns with $-d_{\text{act}}$. These are optimized alongside physical constraints (stability, collision avoidance) and an objective to maximize grasp robustness, measured by the min-weight force closure metric $l^*(q)$~\cite{Ferrari1992, li2023frogger}.
\begin{equation}
\label{eq:grasp_opt_appendix_full} 
{\small
\!\!\begin{aligned}
\max_{q \in \mathcal{Q}} \quad & l^*(q) \\
\text{s.t.} \quad & l^*(q) \ge l_{\min} && \text{(Min F. Closure)}\\ 
                 & s(FK_i(q)) = 0,\ \forall i \in \{1, ..., n_c\} && \text{(Surface Contact)}\\
                 & \sigma_j(q) \ge d_j,\ \forall \text{ collision pairs } j && \text{(Coll. Avoidance)}\\
                 & \exists i_{act} \in F_{act} \text{ s.t.} \\ 
                 & \quad \|FK_{tip}(q, i_{act}) - p_{act}\|_2 \le \delta_{dist} && \text{(Act Pt Proximity)}\\
                 & \quad n_{pad}(q, i_{act}) \cdot (-d_{act}) \ge \cos(\delta_{angle}) && \text{(Act Alignment)} 
\end{aligned}}
\end{equation}
If the optimization fails, we resample $q_0$ and restart. This process yields a diverse set of feasible, function-aligned candidates (Fig.~\ref{fig:grasp_examples}) for initializing the RL policy.

\subsubsection{Constraint-Guided RL}

While the analytic candidates provide functionally-aligned and statically stable starting points, they do not account for the dynamics of the full manipulation task. The goal of this stage is therefore to learn a policy $\pi$ that learns a dynamically robust motion sequence for the entire grasp-move-actuate task. By initializing the RL policy with the optimized candidates, we significantly constrain the exploration problem, enabling the agent to focus on learning the complex dynamics of contact, movement, and actuation.

\textbf{Policy Parametrization} The learned policy $\pi$ takes an observation $o$ consisting of the relative hand pose, candidate finger joints, and the designated actuation finger index. It outputs an action vector $a \in \mathbb{R}^{38}$ that parameterizes a motion primitive by defining targets relative to the input candidate $q_{\text{cand}}$:
\begin{itemize}
    \item {Target Grasp Joint Offsets:} $\Delta j_{\text{grasp}} \in \mathbb{R}^{16}$.
    \item {Target Pre-Grasp Joint Percentages:} $p_{\text{pregrasp}} \in \mathbb{R}^{16}$.
    \item {Residual Hand Pose Offset:} $\Delta T_{\text{hand}} \in \mathbb{R}^{6}$.
\end{itemize}

\begin{figure}[t!]
    \centering
    \includegraphics[width=\linewidth]{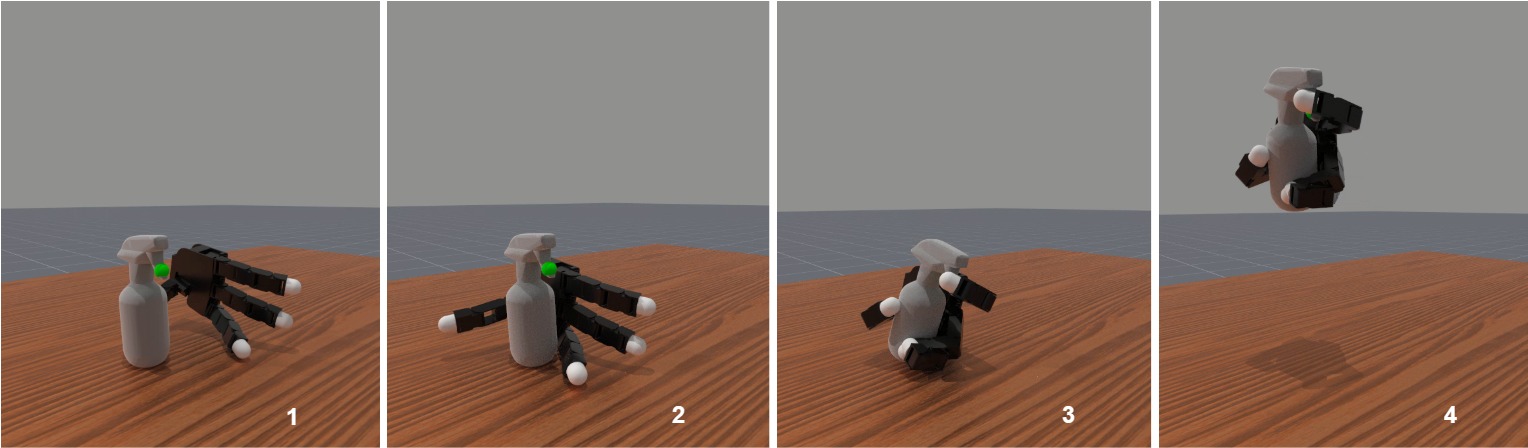} 
    \caption{Key stages of the \methodname{}'s parameterized motion primitive trained in simulation. The policy action space determines (1) the pre-contact approach, (2) grasp pose, (3) finger closing strategy (internal DoF actuation), and (4) object pose change (external DoF actuation).}
    \label{fig:motion_primitive_app}
    \vspace{-1em}
\end{figure}

\textbf{Motion Primitive.} These action parameters guide a multi-stage motion primitive, visualized in Fig.~\ref{fig:motion_primitive_app}. The primitive breaks down the complex task into a sequence of simpler steps: (1) the hand first moves to a pre-contact pose near the object, (2) it then transitions to the final grasp pose, (3) the fingers close to secure the object, (4) finally, it moves the object towards the global goal $T^*$ while simultaneously performing the required actuation. This structured, parameterized primitive makes the high-dimensional control problem tractable for RL.

\textbf{Reward Function} The policy is trained with PPO to maximize a unified reward function $R$. To avoid the need for task-specific reward engineering~\cite{zhang2025dextog}, $R$ is formulated as a normalized weighted sum of continuous shaped rewards ($R_k$), binary stage-completion bonuses ($S_k$), and a final task success bonus ($S_{\text{success}}$). Specifically, the shaped rewards ($R_k$) provide dense feedback for universally required physical behaviors: (1) \textit{grasp stability} (penalizing in-hand object shifting), (2) \textit{lifting} (raising the object while keeping it upright), (3) \textit{actuation alignment} (minimizing the distance between the acting finger and the semantic targets $p_{\text{act}}, d_{\text{act}}$), and (4) \textit{directional force} (applying sufficient force along $d_{\text{act}}$). To heavily discourage dropping the object, a penalty resets the reward to zero if significant grasp displacement is detected. The entire online training process converges in approximately one hour in \textsc{ManiSkill3} simulation~\cite{tao2025maniskill}.

The result of the RL is a full policy that generates a compositional grasp-move-actuate motion for the CD-FOM task.
In the final step, the RL policy is executed on the real robot, using a Franka arm with a LEAP hand.

\section{Experimental Evaluation}
\label{sec:results}

In our experiments, we evaluate whether \textbf{\methodname{}} successfully bridges high-level vision–language understanding and low-level, physics-grounded execution. 
To this end, we investigate the following research questions:
\begin{description}[leftmargin=0.6cm]
\item[\textbf{Q1}] \emph{How well does \textbf{\methodname{}} perform on CD-FOM tasks in the real world?}
\item[\textbf{Q2}] \emph{Do the \textbf{VLM-CEM} global constraints produce semantically and physically valid task poses?}
\item[\textbf{Q3}] \emph{How much does \textbf{constraint-guided RL} improve success over executing grasps from constrained optimization alone?}
\end{description}


\textbf{Experimental Setup:}
We evaluate \methodname{} on a 7‑DoF \textsc{Franka} Emika Panda arm with a 16‑DoF \textsc{LEAP} Hand end-effector, as shown in Fig.~\ref{fig:tasks_objects}. 
Policies are trained in \textsc{ManiSkill3} with 2,048 parallel environments and then directly deployed on the real robot.  
We evaluate six functional manipulation tasks introduced in \S\ref{sec:method}. 
At the start of each episode, the object is placed at a random pose on the table. We evaluate each trial with two binary criteria: (i) correct object movement (the object reaches the task-required pose), and (ii) correct actuation (the mechanism is triggered). A trial is counted as successful only when both criteria are satisfied and no additional failure occurs. For Q3, we also record grasp stability (whether the object slips from the hand during execution). 

\vspace{1ex}
\noindent \textbf{Q1 — Performance on Compositional Dexterous Functional Manipulation Tasks}



\begin{figure}
    \centering
    \includegraphics[width=\linewidth]{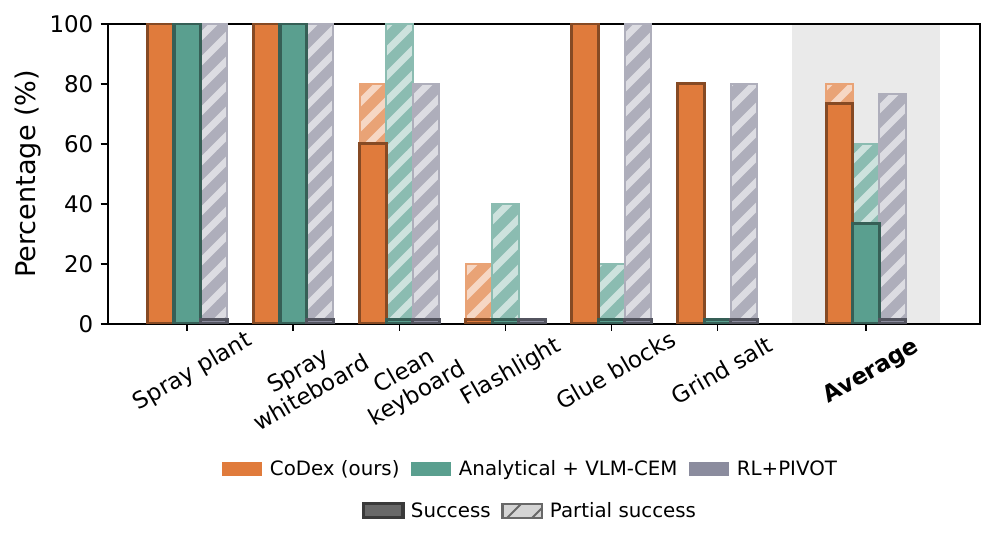}
    \vspace{-2em}
    \caption{Overall success rate comparison on six CD-FOM tasks, evaluated over five real-world trials per task. Solid segments represent the rate of \textbf{success}, while shaded segments show rates of \textbf{partial success} (either correct object movement only or correct internal DoF actuation only). \methodname{} achieves a 73\% success rate, demonstrating the significant benefit of its policy learning stage and VLM-CEM compared to two baselines, pure analytical grasp synthesis~\cite{li2023frogger} and PIVOT~\cite{nasiriany2024pivot}.}
    \label{fig:overall_success}
    \vspace{-1em}
\end{figure}

In real world complete CD-FOM tasks, we compare \methodname{} against two baselines: \textit{Analytical+VLM-CEM} and \textit{RL+PIVOT}. 
The \textit{Analytical+VLM-CEM} baseline combines the best-performing initial grasp candidate from Li et al.~\cite{li2023frogger} (selected via an oracle for maximum stability, see Q3) with the global constraint from our VLM-CEM. The \textit{RL+PIVOT} baseline uses the same policy training procedure as ours but uses PIVOT~\cite{nasiriany2024pivot} (SE3 variant) to generate global constraints.

Figure~\ref{fig:overall_success} summarizes our real-world results.
Our method achieves an overall task success rate of \textbf{73\%}, significantly outperforming the baseline's \textbf{33\%} (Analytical+VLM-CEM) and \textbf{0\%} (RL+PIVOT).  
The performance improvement of our method over {Analytical+VLM-CEM} comes from an increased robustness in the actuation of the dexterous functional manipulation. In the analytical baseline, grasps are often sufficient for lifting and coarse transport, but small in-hand shifts during motion frequently break the precise contact needed for mechanism triggering. 
With our method, thanks to the robustifying effect of policy learning with RL, the robot is able to actuate the internal degrees of freedom after transport and achieves success. We quantify further this reliability gap in Q3.

The baseline RL+PIVOT achieves overall 0\% success rate on entire CD-FOM tasks (although it obtains partial success) because none of PIVOT's generated global constraints meets the task requirements (e.g., the spray bottle is not correctly aimed at the plant or the glue gun is not aimed at the wooden block). 
Although RL+PIVOT can often trigger the object's internal mechanisms, it consistently fails to execute correctly the required object movement. We further analyze this global-constraint mismatch in Q2.  

Failures of \methodname{} primarily arise in two stages of the pipeline: 
1) semantic constraint generation and 2) execution of dexterous functional manipulation.

\textbf{Failures of semantic constraint generation.}
Local constraints are generated reliably and did not cause failures in our experiments. 
However, failures can occur when predicting the global constraint. 
In some cases, the predicted goal pose is spatially incorrect despite appearing plausible in 2D renderings, e.g., in the \texttt{illuminate toy} task the flashlight may appear to aim at the toy in image space while being misaligned in 3D. 
In other cases, the predicted goal pose and the grasp are individually reasonable but jointly infeasible due to joint limits or collisions (e.g., table collision near the target pose in the \texttt{illuminate toy} task). 
Overall, these spatial mismatches and collisions resulted in incorrect object motion in 23.3\% of all trials (87.5\% of all failed trials).

\textbf{Failures of dexterous functional manipulation.}
Successful actuation requires grasps that remain both stable and contact-precise throughout the grasp--move--actuate sequence. 
However, failures arise from sim-to-real discrepancies between training and execution, including geometry reconstructed from 2D images and differences in friction, density, and deformability. 
Across all experiments, actuation failed in 23.3\% of trials (note that some trials include both semantic constraint failures and actuation failures).
The most common failure mode is actuation inaccuracy on high-precision objects. 
This occurs most prominently with the flashlight used in the \texttt{illuminate toy} task, which requires pressing a small deformable button (radius $<0.5\,\mathrm{cm}$) exactly at its center using the robot's relatively large fingers. 
Dexterous functional manipulation of this object failed in all trials. 
Less frequently, sim-to-real discrepancies cause grasp instability, such as occasional salt-grinder toppling or keyboard button slippage.
Stronger domain randomization and closed-loop policies that can reactively correct contact errors are promising directions to mitigate these failures.

\vspace{1ex}
\noindent \textbf{Q2 — Quality of VLM‑CEM Generated Constraints}

While we use binary criteria to assess whether an object's final pose is correct in Q1, some final poses are better than others (e.g., a spray bottle aimed at the edge of the plant versus its center). To evaluate the quality of our VLM-CEM procedure for generating global constraints, we asked twenty participants to rate rendered images generated from the resulting goal poses of VLM-CEM and three additional baselines:

\begin{itemize}[leftmargin=*]
\setlength{\parskip}{0pt}
\setlength{\parsep}{0pt}
\item \textbf{VLM-CEM}: our keypoint-anchored sampler that generates candidate goal poses around detected interaction points on the object.
\item \textbf{VLM-CEM (Dir.)}: a variant that restricts translations sampling to be along the object’s functional axis (e.g., nozzle direction).
\item \textbf{PIVOT (SE3)}~\cite{nasiriany2024pivot}: an adaptation of PIVOT that perturbs full 6-DoF poses in image-space without explicit keypoint anchoring, often resulting in misalignment in depth or lateral offset.
\item \textbf{PIVOT (Trans.)}~\cite{nasiriany2024pivot}: akin to the original PIVOT method, searching only in 2D image-space translations based on coarse visual alignment.
\end{itemize}

We generate three multi-view images per task per method and request human ratings on a five-point scale (1 = unreasonable, 3 = acceptable, 5 = perfect).

\begin{figure}
\centering
\includegraphics[width=\linewidth]{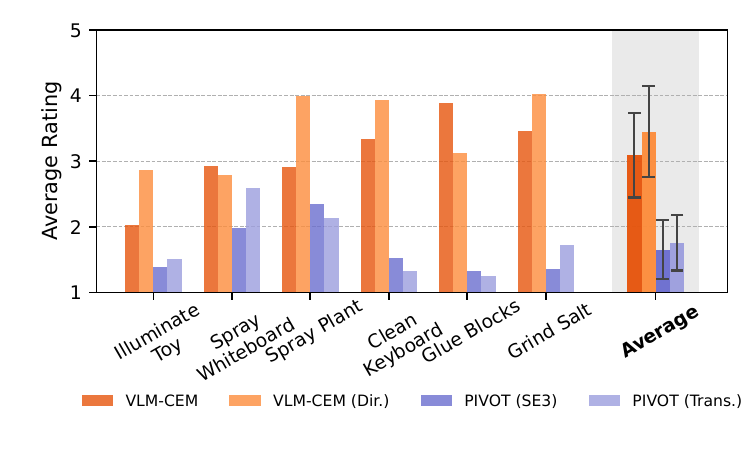}
\vspace{-2em}
\caption{Human study ratings of generated goal poses. We request human feedback on the goal poses generated by our VLM-CEM procedure and baselines (VLM-CEM without rotation changes, PIVOT with rotation and without rotations). We also report the average and standard deviation error bars of the results across all goals for each respective method. On average, the two VLM-CEM methods (ours) are ranked higher in most tasks.}
\label{fig:goal_quality}
\end{figure}

\begin{figure}[t!]
   \centering
   \includegraphics[width=\linewidth]{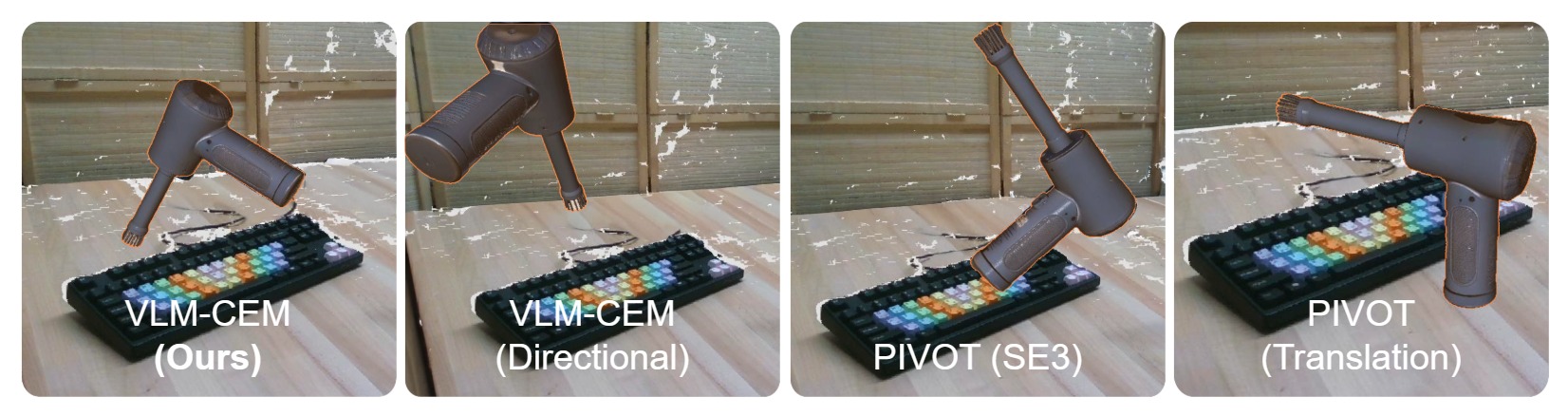} 
   \caption{Example visualizations of different goal-pose-generation methods on the task \texttt{clean keyboard}. Both variants of VLM-CEM generate \textbf{both semantically and physically valid} global constraints, while the baseline methods perform poorly on the task.}
   \label{fig:vlm_cem_more_vis_app} 
   \vspace{-1em}
\end{figure}

Fig.~\ref{fig:goal_quality} shows the human-rating results. Across all tasks, both VLM-CEM variants are ranked above the PIVOT baselines. We verify this trend with a Wilcoxon signed-rank test, obtaining $p<0.02$ for each comparison between the two PIVOT variants and the two VLM-CEM variants.

Surprisingly, for the \texttt{spray plant} task, our VLM-CEM with directional exploration ablation scores significantly higher than the VLM-CEM with full translational exploration. We hypothesize this is because the directional constraint, while reducing the exploration space, effectively filters out candidate poses that might appear plausible in the 2D rendered images used for VLM scoring but are functionally misaligned in 3D (e.g., aiming near the plant but slightly off-axis). By enforcing alignment along the functionally critical nozzle direction, the directional variant ensures better geometric task relevance for the highest-scoring poses in this specific spraying task.

Manual inspection of low-scoring examples supports this interpretation: most come from PIVOT, whose image-space perturbations often introduce lateral or depth offsets relative to the interaction line. In contrast, VLM-CEM samples around detected interaction keypoints, producing goal poses that are both semantically appropriate and geometrically consistent with the task \textbf{(see visual comparisons in Fig.~\ref{fig:vlm_cem_more_vis_app})}.


\vspace{1ex}
\noindent \textbf{Q3 — Benefits of Constraint-Guided RL Policy Learning}

\begin{figure}
\centering
\includegraphics[width=\linewidth]{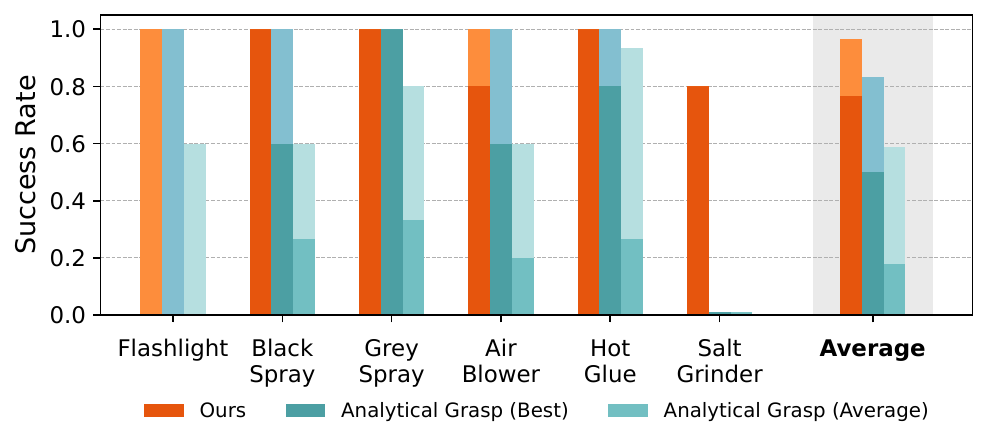}
\vspace{-2em}
\caption{Performance gains of \methodname{} constraint-guided policy training compared to the direct execution of the 3 and the best analytical grasps from \methodname{}'s constrained optimization. Total bar height indicates the success rate of achieving a stable grasp through lifting. The bottom segment (darker shade) represents the success rate of achieving both a stable grasp \textit{and} successful actuation. By training with constraint-guided RL in simulation for the full task, \methodname{} significantly improves stability and actuation of the objects.}
\label{fig:grasp_ablations}
\vspace{-1.2em}
\end{figure}

In this experiment, we isolate the contribution of compositional policy learning by evaluating its improvements in grasp stability and actuation robustness compared to the direct execution of analytical grasps generated via constrained optimization (adapted from \cite{li2023frogger}). For each object, we evaluate three analytical grasp candidates, each repeated five times with randomized object initialization. We report both the average performance across candidates and the \textit{oracle best performance}, i.e., the single best candidate selected post-hoc after evaluating all candidates. Fig.~\ref{fig:grasp_ablations} summarizes the results. 

We observe that our constraint-guided policy learning significantly improves grasp stability, exceeding the average performance of the initial grasp candidates by over 36\% and surpassing the \textit{oracle best performance} (the maximum potential success achievable without refinement) by over 12\%. 
The benefit of policy learning is even more pronounced for functional actuation: the learned policy achieves 60\% higher actuation success than the average grasp candidate and crucially, over 26\% higher success than the best possible outcome using only the initial grasp candidates (oracle).

Interestingly, none of the initial grasp candidates achieved stable grasping success (let alone actuation) for the challenging salt grinder, likely due to its slippery surface and geometry. However, \methodname{}'s policy learning stage successfully discovers a stable grasp, highlighting the method's ability to improve even on difficult cases. 
This demonstrates the importance of the holistic, simulation-based policy learning stage. It allows \methodname{} to refine statically plausible grasp candidates into dynamically robust policies that significantly enhance functional viability compared to executing the initial grasp candidates directly, even when considering the best possible initial grasp candidate.

This stage-wise diagnosis in Q1 is consistent with Q2 and Q3: Q2 isolates the quality of global constraints, while Q3 quantifies how policy learning improves contact maintenance and actuation robustness over direct analytical execution.

\section{Limitations}
\label{sec:limitations}

While \methodname{} demonstrates success across multiple compositional FOM tasks, it also reveals several limitations that we plan to address in future work. First, tasks requiring pinpoint contact, such as pressing tiny push-buttons, are sensitive to finger size and actuator tolerances, demanding higher accuracy in control and possibly different hand morphology to ensure reliable execution. Second, many tasks go beyond reaching a single goal pose and instead require sustained, coupled arm–hand motion—for example, actuating scissors while sliding along paper—calling for extensions toward trajectory-level constraints and closed-loop feedback. Finally, the current policy assumes a single actuation point, leaving out objects that need alternating or multi-point actuation; extending \methodname{} to these tasks will open new robot capabilities.

\section{Conclusion}
\label{sec:conclusion} We addressed the challenging problem of zero-demonstration functional object manipulation by introducing \methodname{}, a framework that translates abstract VLM guidance into concrete semantic constraints. Our method enforces these VLM-generated local (e.g., actuation points) and global (e.g., target poses) constraints through a two-phase process: first, analytic constrained optimization efficiently generates a set of stable, function-aligned grasp candidates, and second, constraint-guided reinforcement learning initializes from these candidates to discover the complete, dynamically robust grasp-move-actuate policy. Our experiments on a physical robot demonstrated that this tight integration of semantic reasoning with a physics-grounded, constraint-enforcing pipeline is crucial. \methodname{} autonomously discovered and executed complex strategies for diverse tasks with unseen objects, validating our approach. This work represents a key step toward versatile and autonomous tool manipulation. Future directions include extending the framework to more complex object mechanisms, incorporating tactile feedback for finer control, and exploring richer, interactive VLM dialogues.

\bibliographystyle{IEEEtran}
\bibliography{bibliography}

\clearpage

\end{document}